\documentclass[letterpaper]{article} 
\usepackage{aaai24}  
\usepackage{times}  
\usepackage{helvet}  
\usepackage{courier}  
\usepackage[hyphens]{url}  
\usepackage{graphicx} 
\usepackage{natbib}  
\usepackage{caption} 
\usepackage{algorithm}
\usepackage{algorithmic}

\usepackage{amsmath,amsfonts}
\usepackage{amsthm,amsmath,amssymb}

\usepackage{mathrsfs}
\usepackage{bm}

\usepackage{multirow} 
\usepackage{array}
\usepackage{booktabs}
\usepackage[caption=false,font=normalsize,labelfont=sf,textfont=sf]{subfig}
\usepackage{textcomp}
\usepackage{url}
\usepackage{verbatim}
\usepackage{cite}
\usepackage[font=small]{caption}
\usepackage{newfloat}
\usepackage{listings}
\DeclareCaptionStyle{ruled}{labelfont=normalfont,labelsep=colon,strut=off} 
\floatstyle{ruled}
\newfloat{listing}{tb}{lst}{}
\floatname{listing}{Listing}
\title{Multi-modality Affinity Inference for Weakly Supervised 3D Semantic Segmentation}
\author {
    Xiawei Li\textsuperscript{\rm 1}\equalcontrib,
    Qingyuan Xu\textsuperscript{\rm 1}\equalcontrib,
    Jing Zhang\textsuperscript{\rm 1}\thanks{Corresponding author},
    Tianyi Zhang\textsuperscript{\rm 2},
    Qian Yu\textsuperscript{\rm 1},
    Lu Sheng\textsuperscript{\rm 1},
    Dong Xu\textsuperscript{\rm 3}
}
\affiliations {
    \textsuperscript{\rm 1}School of Software, Beihang University\\
    \textsuperscript{\rm 2}College of Computer Science and Technology, Zhejiang University\\
    \textsuperscript{\rm 3}Department of Computer Science, The University of Hong Kong\\
    \{ZY2121108,ZY2121121,zhang\_jing,qianyu,lsheng\}@buaa.edu.cn, 
    tianyizhang0213@zju.edu.cn, dongxu@hku.hk
}

\usepackage{bibentry}

\begin{document}

\maketitle

\begin{abstract}
3D point cloud semantic segmentation has a wide range of applications. Recently, weakly supervised point cloud segmentation methods have been proposed, aiming to alleviate the expensive and laborious manual annotation process by leveraging scene-level labels. However, these methods have not effectively exploited the rich geometric information (such as shape and scale) and appearance information (such as color and texture) present in RGB-D scans. Furthermore, current approaches fail to fully leverage the point affinity that can be inferred from the feature extraction network, which is crucial for learning from weak scene-level labels. Additionally, previous work overlooks the detrimental effects of the long-tailed distribution of point cloud data in weakly supervised 3D semantic segmentation. To this end, this paper proposes a simple yet effective scene-level weakly supervised point cloud segmentation method with a newly introduced multi-modality point affinity inference module. The point affinity proposed in this paper is characterized by features from multiple modalities (e.g., point cloud and RGB), and is further refined by normalizing the classifier weights to alleviate the detrimental effects of long-tailed distribution without the need of the prior of category distribution. Extensive experiments on the ScanNet and S3DIS benchmarks verify the effectiveness of our proposed method, which outperforms the state-of-the-art by $\sim 4\%$ to $\sim 6\%$ mIoU. Codes are released at \url{https://github.com/Sunny599/AAAI24-3DWSSG-MMA}.

\end{abstract}

\section{Introduction}

Point cloud data capture rich object and scene geometric and appearance information, which serves as an essential data representation for various applications, such as autonomous driving, augmented reality, and robotic. Point cloud semantic segmentation plays a key role in 3D scene understanding, and has been extensively explored~\cite{qi2017pointnet,qi2017pointnet++,wang2019dynamic,wang2019graph,thomas2019kpconv,shi2019pointrcnn,wu2019pointconv,hu2020randla,zhao2021point,xu2021paconv,lai2022stratified}. However, the success of most of the methods is based on learning in a fully supervised manner, requiring extensive point-level annotations. 

To reduce the annotation costs, some recent works have delved into the realm of weakly supervised semantic segmentation (WSSS) methods~\cite{Bearman2016,xu2020weakly,Zhang2021,liu2021one,hou2021exploring,zhang2021weakly,yang2022mil,wei2020multi,ren20213d,yang2022mil}. These methods can be categorized based on different levels of supervision, including partially labeled points, sub-cloud level annotations, and scene-level annotations. Among these, segmentation with scene-level labels is the most challenging scenario as point-wise annotations are completely unavailable, which is the focus of our paper.
 

Most of the current methods with scene-level annotations~\cite{wei2020multi,ren20213d,yang2022mil} are proposed based on pseudo labels which can be obtained by Class Activation Map (CAM)~\cite{zhou2016learning} or Multiple Instance Learning (MIL)~\cite{Maron1997}. The key to successful weakly supervised semantic segmentation is how to expand the semantic regions to achieve completeness and preciseness of the localized objects. The 3D point clouds of the RGB-D scans provide accurate object shape and scale information without occlusion and distortions, while the corresponding RGB data provide additional color and texture information. However, the current weakly supervised point cloud segmentation based on RGB-D scans fails to fully take advantage of all the data modalities. Moreover, current methods fail to fully exploit the point affinity that can be readily inferred from the feature extraction network. We argue that point affinity that characterizes the similarity between points is essential for learning from scene-level weak labels. 

In addition, the long-tailed distribution of point cloud data due to the extremely imbalanced point data have detrimental effects on both point-wise classification performance and point-wise feature learning, which has largely overlooked by previous research on weakly-supervised point cloud semantic segmentation. For example, when the point cloud data have long-tailed distribution, the point-wise classifier for semantic segmentation will be biased towards the head classes with more number of samples \cite{9739747,kang2019decoupling,wang2021adaptive,cui2019class,jamal2020rethinking}. Moreover, the point features of the tail classes with less number of samples will also be learned similar to the point features of the head classes since the feature extractor may also be dominated by the head classes. These detrimental effects will further affect the affinity learning that relies on feature similarity. However, the lack of per-point category information in scene-level weak supervision makes it challenging to obtain category distribution. Therefore, how to take full advantage of both RGB and geometry information in the point cloud data while addressing the long-tailed distribution issue for affinity learning in the context of weakly-supervised semantic segmentation remain a challenging task.

To this end, we propose a novel multi-modality affinity (MMA) enhanced weakly supervised semantic segmentation (WSSS) method by fully exploiting the point feature affinities from multiple data modalities and eliminating the influence of long-tail data distribution. The geometric information derived from point clouds and the color and texture information captured in RGB data offer distinct perspectives for characterizing feature affinity. By leveraging these complementary data modalities, we propose to generate multi-modality affinities based on both pure geometric data as well as color-appended RGB-D data. For simplicity, we mask out the RGB features from the input point cloud data to model the geometric affinity and use the original point cloud data to model the RGB-enriched data affinity. We also enhance the affinity by normalizing the weights of the point-wise classifier. This normalization process assists in mitigating the network's tendency to misclassify data points into the dominant head categories. Consequently, it enhances the point-wise features of the tail categories, allowing for improved affinity inference. The obtained multi-modality affinity matrices are used to refine the WSSS-related objective functions. Extensive experiments are conducted on ScanNet~\cite{dai2017scannet} and S3DIS~\cite{armeni20163d} benchmarks, and the results demonstrate that our method significantly outperforms the state-of-the-art scene-level weakly supervised point cloud semantic segmentation methods.



\begin{figure*}
    \centering
   \includegraphics[width=12cm]{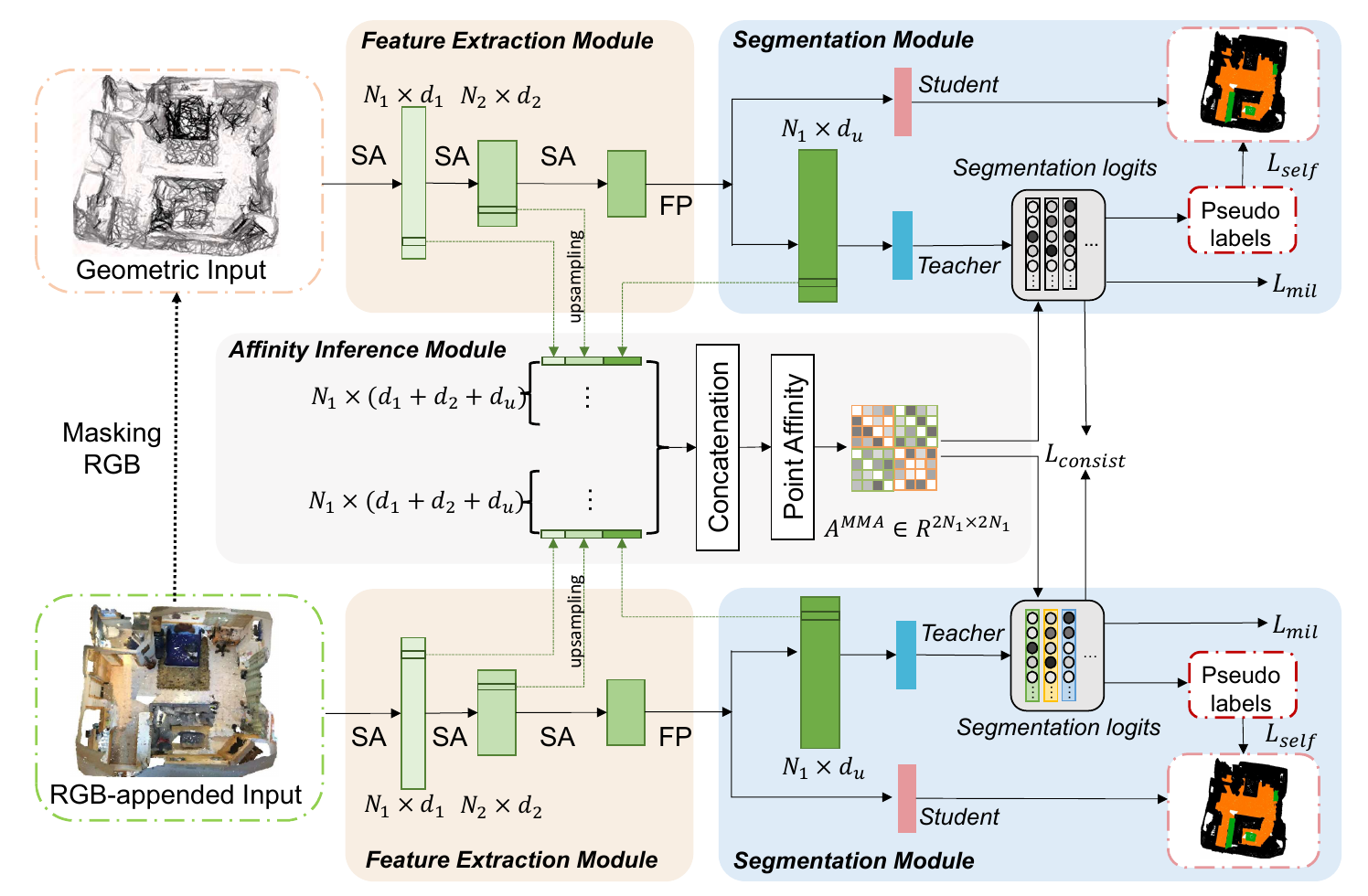}
    \caption{Pipeline of the proposed method. A two-stream architecture with shared parameters is adopted, where the two streams take pure geometric point clouds and RGB-appended point clouds as inputs, respectively. Three main modules are involved: a \textit{feature extraction module}, a \textit{segmentation module}, and a \textit{multi-modality affinity inference module}. }
    \label{fig:galaxy}
    \vspace{-1em}
\end{figure*}

\section{Related Work}

\subsection{Weakly-supervised Point Cloud Segmentation with Sparse Labels.}The main idea of learning from sparse annotations focuses on propagating information from labeled points to the unlabeled points. For example, Liu \textit{et al.} \cite{liu2021one} generates a super-voxel graph between labeled points and unlabeled to guide the iterative training. Yang \textit{et al.}\cite{yang2022mil} design transformer model derived by multiple instance learning (MIL), where the two clouds with shared category yield a positive bag while with different classes produce a negative bag. 
Recently, PSD~\cite{Zhang2021} proposes to learn 3D point affinity based on sparsely labeled points. Though the affinity can be precisely learned based on the supervised learning loss with sparsely labeled points, this method is not directly applicable to our setting, where only the scene-level category supervision is provided without any ground-truth point-level supervision.
Therefore, these methods with sparse labels mainly focus on relationship between labeled and unlabeled points, which is not applicable in our setting when point-level supervision is not available.

\subsection{Weakly-supervised Point Cloud Segmentation with Scene-level Labels.}
Compared to weakly supervised point cloud segmentation based on sparse point-level annotations, the methods by scene-level annotations are less exploited. The state-of-the-art methods generally generate pseudo labels in the first step, and then refine the segmentation results via self-training.
Two types of strategies are commonly used for generating pseudo labels by previous methods: class activation maps (CAM)-based and multi-instance learning (MIL)-based. Wei \textit{et al.}\cite{wei2020multi} propose a Multi-Path Region Mining model involving spatial, channel, and point-wise paths to generate CAM of sub-cloud as pseudo. It is common to design two branches for a network and use consistency loss for self-supervised learning between the original point cloud and the augmented\cite{ren20213d} or perturbed \cite{zhang2021weakly} point clouds. Similarly, Ren \textit{et al.}\cite{ren20213d} propose to jointly learn semantic segmentation, 3D proposal generation, and 3D object detection in a two-branch framework. 
However, current methods do not effectively leverage the complementary RGB and geometric information present in point cloud data. Additionally, these methods often overlook the long-tail distribution of different categories, resulting in suboptimal performance.

\section{Methodology}

Given a set of $M$ point clouds with scene-level annotations: $D=\{P_m,\bm{y}_m\}_{m=1}^M$, where $P_m\in{\mathbb{R}^{N\times (3+K)}}$ denotes the $m$th point cloud and $\bm{y}_m\in \{0,1\}^C$ is a $C$-dimensional binary vector indicating which categories are present in point cloud $P_m$, we aim to derive a segmentation model, which classifies each point into one of the $C$ categories. Each $P_m\in{\mathbb{R}^{N\times (3+K)}}$ represents a whole 3D scene with $N$ 3D coordinates together with K-dimensional auxiliary features, such RGB, object normals, and height. 

\paragraph{Overview.} As shown in Figure~\ref{fig:galaxy}, our framework consists of three main modules: a \textit{feature extraction module} and a \textit{segmentation module}, and a \textit{multi-modality affinity inference module} to enhance the segmentation results.
Specifically, our \textbf{feature extraction module} hierarchically extract multiple scales of point set features by gradually grouping point features in local regions. In the \textbf{multi-modality affinity inference module},
to differentiate between pure geometric data and RGB-enriched data, we employ a masking technique to exclude the RGB features from the input data, resulting in the pure geometric data. Simultaneously, we retain the original point cloud data as the RGB-enriched data.
The two types of input data are fed into the shared backbone network to obtain the respective point affinity of the corresponding modality. The \textbf{segmentation module} then utilizes the learned point affinity to refine the MIL (Multiple Instance Learning) objective, as well as the point-level pseudo-labels. These refinements are instrumental in guiding the self-training process of semantic segmentation.

\subsection{MIL-based 3D Semantic Segmentation}
\label{sec:baseline}
We use a MIL-based weakly-supervised 3D semantic segmentation model as our baseline model, which consists of a feature extraction network and a segmentation head.
\paragraph{Feature Extraction Module.}
In this work, we choose our backbone network based on PointNet++~\cite{qi2017pointnet++} to hierarchically extract multiple scales of point set features by gradually grouping point features in local regions. The four set abstraction (SA) layers down-sample the point cloud from $N$ points to $N_1=2048$, $N_2=1024$, $N_3=512$, and $N_4=256$ points, respectively by each layer. Two feature propagation (FP) layers then up-sample the points from $N_4$ to $N_2$ points. Note that other backbone networks that extract multiple scales of point features can also be selected as the feature extraction network.
\paragraph{Segmentation Module.}
Inspired by~\cite{ren20213d}, we use two segmentation heads to process extracted features from the backbone networks. Each head predicts a segmentation logits matrix over $C$ classes for all points, denoted as $U_{seg}\in \mathbb{R}^{N_1\times C}$ and $S_{seg}\in \mathbb{R}^{N_1\times C}$, respectively. The first head is to find the semantically discriminative points through a MIL-loss and produce pseudo labels by the points with highly confident predictions, which is denoted as the teacher head. The pseudo labels are generated by following~\cite{ren20213d}. The second head takes the pseudo labels produced by the first head for self-training, which is denoted as the student head. However, we use a slightly different structure for the teacher and student segmentation heads. Specifically, the teacher head contain one FP layer to up-sample the points from $N_2=1024$ to $N_1=2048$ and one fully-connected (FC) layer to predict the per-point segmentation logits. 
The student head contains one FP layer and is followed by a one-layer Transformer Encoder~\cite{vaswani2017attention}. The motivation for adding a Transformer layer to the student head is two-fold.
The first is to borrow the self-attention module in Transformer for capturing long-range dependencies to ensure object completeness. The second is to explicitly enforce the teacher head and the student head to learn different features such that the pseudo labels produced by the teacher head could benefit more on the student head, which is inspired by co-training~\cite{blum1998combining}.
\paragraph{Loss Function.}
The loss function of our weakly supervised semantic segmentation module is defined as,
\begin{equation}
\mathcal{L}_{wsss}=\mathcal{L}_{mil}+\mathcal{L}_{self},
    \label{eq:wsss}
\end{equation}
where $\mathcal{L}_{mil}$ is the MIL loss and $\mathcal{L}_{self}$ is the self-training loss. We define the scene-level MIL loss as,
\begin{equation}
\small
    \mathcal{L}_{mil}=-\sum_{c=1}^C (\bm{y}[c] \log \bm{\sigma} [c]-(1-\bm{y}[c]) log(1-\bm{\sigma}[c])),
    \label{eq:mil}
\end{equation}
where 
    $\bm{\sigma}[c]=sigmoid(\frac{1}{N_1}\sum_{i=1}^{N_1} U_{seg}[i,c])$
converts the per-point logits $U_{seg}$ into a scene-level prediction $\bm{\sigma}$ via average pooling and a sigmoid activation.
The self-training loss for each scene is formulated by
\begin{equation}
\small
    \mathcal{L}_{self}=-\frac{1}{N_1}\sum_{i=1 }^{N_1}\sum_{c=1}^C \hat{Y}[i,c] \log \psi [i,c],
    \label{eq:self}
\end{equation}
where $\psi [i,c]=softmax(S_{seg}[i,c])$ denotes the probability of the $i$th point being predicted as class $c$, $N_1$ denotes the number of points in $U_{seg}$,  and $\hat{Y}[i,c]\in \{0,1\}$ is the point-wise pseudo label generated by the teacher head~\cite{ren20213d} by selecting the high confident points within the scene categories.
\subsection{Long-tail-aware Multi-modality Point Affinity}

\paragraph{Multi-modality Point Affinity.}

We argue that geometric information from point cloud and color information encoded in RGB data can characterize the feature affinity from different perspectives by taking advantages of different data modalities. Moreover, to make the most of the multi-scale features extracted by the backbone, we incorporate features from multiple layers of feature extractor. We obtain the RGB-appended point affinity by concatenating multiple scales of point features $F\in{\mathbb{R}^{N_1\times D}}=concate(F_1, F_2, F_u)$, where $F_1\in{\mathbb{R}^{N_1\times d_1}}, F_2\in{\mathbb{R}^{N_1\times d_2}}, F_u\in{\mathbb{R}^{N_1\times d_u}}$. $F_u$ is obtained by teacher head, is a high-level abstract points representation. $F_1$ and $F_2$ indicate the multiple scales of features produced by different SA layers from the backbone network. Note that the $N_1$ points of $F_2$ is obtained by up-sampling via linear interpolation. We model the geometric affinity by masking out the RGB values from the input point cloud for simplicity.
Similarly, we use $\Tilde{F}\in{\mathbb{R}^{N_1\times D}}=concate(\Tilde{F}_1, \Tilde{F}_2, \Tilde{F}_u)$ to denote the multi-scale geometric features by aggregating multiple scales of features from RGB-masked point clouds. We define the multi-scale multi-modality features by concatenating the multi-scale geometric features and RGB-appended features as $F^{M}=[F\in{\mathbb{R}^{N_1\times D}};\Tilde{F}\in{\mathbb{R}^{N_1\times D}}]$ and $F^{M}\in{\mathbb{R}^{2N_1\times D}}$.
Thus, our final multi-scale multi-modality affinity matrix is defined as,
\begin{equation}
\small
 A^{MMA}[i,j]=\max\left(\theta, \frac{\langle F^{M}[i,\cdot],F^{M}[j,\cdot]\rangle}{\|F^{M}[i,\cdot]\|\|F^{M}[j,\cdot]\|}\right),
  \label{eq:MMA-affinity}
\end{equation}
where $i,j=1,2,...,2N_1$, and $\theta$ is the threshold for filtering out less confident affinities. Since the multi-scale multi-modality affinity matrix $A^{MMA}\in{\mathbb{R}^{2N_1\times 2N_1}}$ is defined based on features from multiple modalities, the similarity of point features across-modality is also considered in our affinity matrix.

\paragraph{Long-tail Aware Affinity Enhancement.}

In point cloud-based semantic segmentation, the long-tailed distribution of data among classes happens at both the category level and point level. For example, in indoor scenarios, the category ``wall'' and ``floor'' generally not only appear in almost all of the scenes but also contain a larger amount of points in each scene compared to other objects, which are termed head classes in long-tailed distribution. 
The long-tailed distribution leads to the classifier weights of the head classes are learned to have larger norms, resulting in greater logits for head classes in each sample. This can be attributed to that large classifier weights norms cause the gradient leans towards head classes during back propagation. This biases the network's learned features towards head categories. Since the affinity matrix is calculated based on feature similarity, the head classes data points will contribute more to the affinity values than the tail classes, which contradict to the fact that the within-class point features should have larger affinity values.
Therefore, the long-tail issue not only affects the final segmentation results but also is detrimental to affinity inference. Unfortunately, previous research has overlooked this issue, and utilizing affinity affected by the long-tail distribution may lead to error accumulation.

In WSSS, the point-level category distribution is not accessible with scene-level annotations. Thus, inspired by Decoupling Representation and Classifier~\cite{kang2019decoupling}, we enhance affinity inference by dealing with the long-tail issue via a simple method through normalizing classifier weight(NCW), which alleviates the long-tail issue without the need of the prior of category distribution. Formally, let $W=\{\bm {w_i}\}\in \mathbb{R}^{d\times C}$, where $\bm {w_i}\in \mathbb{R}^d$ are the classifier weights corresponding to class $i$ of the teacher and student segmentation head. We normalize $W$ to obtain $\widehat{W}=\{\widehat{\bm {w_i}}$\} via $\widehat{\bm {w_i}}=\frac{\bm {w_i}}{\|\bm {w_i}\|}$,
where $\| \cdot\|$ denotes the $l_2$ norm.
\subsection{Objective Functions}
With the learned multi-modality affinity matrix $A^{MMA}$, we obtain the refined segmentation logits matrix $U_{seg}^{refined}$ and $ \Tilde{U}_{seg}^{refined}$ of the teacher segmentation head by multiplying the original logits matrix by the affinity matrix:
\begin{equation}
\small
\left[U_{seg}^{refined};\Tilde{U}_{seg}^{refined}\right]=A^{MMA} \left[U_{seg};\Tilde{U}_{seg}\right],
\end{equation}
where $\Tilde{U}_{seg}\in \mathbb{R}^{N_1\times C}$ denotes the predicted segmentation logits matrix over $C$ classes of the RGB-masked point cloud data $\Tilde{P}$.
The refined segmentation logits of a point aggregate information from the points with similar features both within and across modalities, which are considered to be able to improve the pseudo labels for self-training loss in Eq.(\ref{eq:self}).

To achieve message passing between multi-modality affinities and further improve the segmentation performance, 
we explicitly impose the prediction consistency constraint between the original point cloud data $P$ and the RGB-masked point cloud data $\Tilde{P}$ with horizontal transformation, which is inspired from previous WSSS methods that introduce different contrastive or consistency learning strategies by augmenting the original point cloud data~\cite{yang2022mil,ren20213d,ahn2018learning,Zhang2021}. The consistency loss is defined as,
\begin{equation}
\small 
\mathcal{L}_{consist}=\frac{1}{N_1}\sum_{i=1 }^{N_1}\sum_{c=1}^C \left|U_{seg}^{refined}[i,c]-\Tilde{U}_{seg}^{refined}[i,c]\right|.
  \label{eq:consist}
\end{equation}
Moreover, the refined self-training loss $\mathcal{L}_{self}^{refined}$ is defined:
\begin{equation}
\small
    \mathcal{L}_{self}^{refined}=-\frac{1}{N_1}\sum_{i=1 }^{N_1}\sum_{c=1}^C \hat{Y}^{refined}[i,c] \log \psi [i,c],
    \label{eq:self_refined}
\end{equation}
where $\hat{Y}^{refined}[i,c]\in \{0,1\}$ is the point-wise pseudo label generated by $U_{seg}^{refined}$.

The final objective function of our method is
\begin{equation}
\small
    \mathcal{L} = \mathcal{L}_{mil}+    \mathcal{L}_{self}^{refined}+ \mathcal{L}_{consist}.
\end{equation}

\section{Experiments}

\subsection{Experimental setting}
\paragraph{Datasets and evaluation metrics.}
We evaluate the proposed approach MMA on two benchmarks, ScanNet~\cite{dai2017scannet} and S3DIS~\cite{armeni2017joint} datasets. ScanNet is a commonly-used indoor 3D point cloud dataset for semantic segmentation. It contains 1513 training scenes (1201 scenes for training, 312 scenes for validation) and 100 test scenes,  annotated with 20 classes. S3DIS is also an indoor 3D point cloud dataset, which contains 6 indoor areas and has 13 classes. By following the previous work, we use area 5 as the test data. Mean intersection over union (mIoU) is used as the evaluation metric of the segmentation results. 
\paragraph{Implementation details.}
The RGB-appended input point clouds are a set of 10-dimensional vectors, including coordinates (x,y,z), color (R,G,B), surface normal, and height, while the pure geometric input is produced by masking out the RGB values with 0. 
PointNet++~\cite{qi2017pointnet} is adopted as the backbone feature extraction module to extract point cloud features. 
The segmentation module has two segmentation heads: a teacher head for providing pseudo labels with MIL loss and a student head for self-training. 
The teacher head is a multi-label classification model, which contains one FP layer to upsample the points to 2048 and one fully-connected (FC)  layer to predict per-point logit and then through average to get per-class logit.
The student head contains one FP layer to upsample points and follows one Transformer Encoder layer to capture the long-range dependencies of points, then an FC layer to predict per-point logit. Multi-scale module use features obtained from the first and second SA layers and $U_{seg}$ FP layer. The details of the architecture of teacher head and student head can be found in the supplementary material.
The model is trained on 3090 GPU with batch size 8 for 300 epochs. We use AdamW optimizer with an initial learning rate of 0.0014 and decay to half at 160 epochs and 180 epochs. All hyper-parameters are tuned based on the validation set. 


\subsection{Comparison with State-of-the-arts}
We mainly compare our approach to other 3D weakly supervised segmentation methods utilizing scene labels, including MPRM~\cite{wei2020multi}, WyPR~\cite{ren20213d}, and MIL-Derived~\cite{yang2022mil}. This type of supervision is challenging for large-scale point cloud datasets. MPRM uses various attention modules to mine local and global context information. WyPR joint learning of segmentation and detection to get a better feature representation, and gain high performance. MIL-Derived proposes a transformer model to explore pair-wise cloud-level supervision, where two clouds of the same category yield a positive bag while two of different classes produce a negative bag.

\begin{table}[!ht]
    \centering
    \caption{3D semantic segmentation on ScanNet.}
    \vspace{-1em}\resizebox{\linewidth}{!}{\begin{tabular}{c|c|c|c}
    \hline
        Method & Supervision & Val. & Test  \\ \hline
        PointNet++~\cite{qi2017pointnet++}  & Full & - & 33.9  \\
        PointCNN~\cite{li2018pointcnn}  & Full & - & 45.8  \\
        KPConv~\cite{thomas2019kpconv} & Full & - & 68.4  \\ 
        MinkNet~\cite{choy20194d}  & Full & - & 73.6  \\ \hline
        MPRM~\cite{wei2020multi} & Scene & 21.9 &  - \\ 
        WyPR~\cite{ren20213d} & Scene & 29.6 & 24  \\ 
        WyPR+prior~\cite{ren20213d} & Scene & 31.1 & -  \\ 
        MIL-Derived~\cite{yang2022mil} & Scene & 26.2 &  - \\ 
        Ours (MMA) & Scene & \textbf{37.7} & \textbf{30.6}  \\ \hline
    \end{tabular}}
    \label{tab:scannet}
    \vspace{-1em}
\end{table}

\paragraph{Results on ScanNet.}
Table~\ref{tab:scannet} reports the mIoU results of the proposed method and the state-of-the-art baseline methods. It can be seen that our proposed method performs better than existing methods (MPRM~\cite{wei2020multi}, WyPR~\cite{ren20213d}, MIL-Derived~\cite{yang2022mil})  by large margins(+15.8\%, +8.1\%, +11.5\%) on the ScanNet validation set. And our method outperforms WyPR by 6.6\% in terms of the test mIoU. In Table~\ref{tab:scannet-perclass}, we report the per-class IoU on ScanNet. Obviously, the proposed MMA achieves the highest mIoU, and significantly improves the performance in ``floor'',  ``chair'', ``sofa'', ``table'', ``shelf'', ``desk'', ``toilet'' and ``bathtub'' against WyPR. These categories are often co-occurred and easily mis-classified.
In addition, our ``MMA'' as shown in Table \ref{tab:scannet-perclass} improves the performance of objects with either discriminative geometric or color information, such as ``floor'', ``chair'', ``table'', ``curtain'', ``shower curtain'', and ``bathtub''. The Multi-modality affinity in MMA takes advantage of both geometric and color information and thus improves the performance to a large margin.

\begin{table}[!h]
\vspace{-1em}
    \centering
    \small
    \caption{{3D semantic segmentation on S3DIS.}}
    \label{tab:s3dis}
    \vspace{-1em}
    \resizebox{\linewidth}{!}{\begin{tabular}{c|c|c}
    \hline
        Method & Supervision & Test  \\ \hline
        PointNet++~\cite{qi2017pointnet++}  & Full & 53.5  \\
        PointCNN~\cite{li2018pointcnn}  & Full & 57.3 \\
        KPConv~\cite{thomas2019kpconv} & Full & 70.6  \\ 
        MinkNet~\cite{choy20194d} & Full & 65.4  \\ \hline
        MPRM~\cite{wei2020multi} & Scene & 10.3  \\ 
        WyPR~\cite{ren20213d} & Scene & 22.3  \\ 
        MIL-Derived~\cite{yang2022mil} & Scene & 12.9   \\ 
        Ours (MMA) & Scene & \textbf{26.3}  \\ \hline
    \end{tabular}}
    \vspace{-1em}
\end{table}

\begin{table*}[t]
    \centering
    \caption{3D semantic segmentation result for 20 classes on ScanNet. Here sc refers to the `shower curtain' class}
    \vspace{-1em}
    \resizebox{\linewidth}{!}{
    \begin{tabular}{c|c|ccccccccccccccccccccc}
        \toprule
        Method & eval. & mIoU & wall & floor & cabinet & bed & chair & sofa & table & door & window & shelf & picture & counter & desk & curtain & fridge & sc & toilet & sink & bathtub & other  \\ 
        \midrule
        PCAM~\cite{wei2020multi} & train & 22.1 & 54.9 & 48.3 & \textbf{14.1} & 34.7 & 32.9 & 45.3 & 26.1 & 0.6 & 3.3 & 46.5 & 0.6 & 6.0 & 7.4 & 26.9 & 0 & 6.1 & 22.3 & 8.2 & 52.0 & 6.1 \\
        MPRM~\cite{wei2020multi} & train & 24.4 & 47.3 & 41.1 & 10.4 & 43.2 & 25.2 & 43.1 & 21.5 & 9.8 & 12.3 & 45.0 & 9.0 & 13.9 & 21.1 & 40.9 & 1.8 & 29.4 & 14.3 & 9.2 & 39.9 & 10.0  \\ 
        WyPR~\cite{ren20213d} & train & 30.7 & \textbf{59.3} & 31.5 & 6.4 & 58.3 & 31.6 & 47.5 & 18.3 & 17.9 & \textbf{36.7} & 34.1 & 6.2 & \textbf{36.1} & 24.3 & \textbf{67.2} & \textbf{8.7} & 38 & 17.9 & \textbf{28.9} & 35.9 & 8.2  \\ 
        Ours(MMA) & train & \textbf{41.2} & 52 & \textbf{84.2} & 13.2 & \textbf{63.2} & \textbf{53.1} & \textbf{62.4} & \textbf{41.1} & \textbf{18.0} & 31.1 & \textbf{46.6} & \textbf{10.2} & 24.6 & \textbf{43} & 65.8 & 6.6 & \textbf{59.8} & \textbf{43} & 24.3 & \textbf{64.3} & \textbf{17.6}  \\
        \midrule
        WyPR~\cite{ren20213d} & val & 29.6 & \textbf{58.1} & 33.9 & 5.6 & 56.6 & 29.1 & 45.5 & 19.3 & 15.2 & \textbf{34.2} & 33.7 & 6.8 & \textbf{33.3} & 22.1 & 65.6 & 6.6 & 36.3 & 18.6 & 24.5 & 39.8 & 6.6  \\ 
        WyPR+prior~\cite{ren20213d} & val & 31.1 & 52.0 & 77.1 & 6.6 & 54.3 & 35.2 & 40.9 & 29.6 & 9.3 & 28.7 & 33.3 & 4.8 & 26.6 & 27.9 & \textbf{69.4} & \textbf{8.1} & 27.9 & 24.1 & 25.4 & 32.3 & 8.7  \\ 
        Ours(MMA) & val & \textbf{37.7} & 49.8 & \textbf{84.3} & \textbf{11.1} & \textbf{56.8} & \textbf{54.3} & \textbf{51.6} & \textbf{45.2} & \textbf{16.0} & 22.2 & \textbf{42.0} & \textbf{6.8} & 28.6 & \textbf{37.2} & 59.9 & 4.2 & \textbf{43.0} & \textbf{41.6} & \textbf{27.4} & \textbf{59.7} & \textbf{12.3}  \\ 
        \bottomrule
    \end{tabular}
    }
    \label{tab:scannet-perclass}
    \vspace{-1.5em}
\end{table*}
\paragraph{Results on S3DIS.} Table~\ref{tab:s3dis} shows the S3DIS results of the proposed method and the baseline methods. It can be seen that our proposed method achieves much higher mIoU scores, and outperforms MPRM, MIL-Derived and WyPR with gains of 16.0\%, 13.4\% and 4.0\%. 

\vspace{-1em}
\hspace*{\fill}
\paragraph{Discussion.} When compared with the state-of-the-art methods, our method owns different advantages. Firstly, MPRM~\cite{wei2020multi} proposes various attention modules to refine the point features by local and global context information. The attention modules are only applied to the output features of the backbone network, which might fail to capture multi-scale attentions. Moreover, the multiple modalities of RGB-D data are not explicitly exploited. By contrast, our MMA affinity takes both multi-scale and multi-modality similarities into account and the obtained MMA affinity matrix is directly applied to the segmentation logits the refine the segmentation results, which directly improve the pseudo labels for self-training. Secondly, WyPR~\cite{ren20213d} achieves better results than MPRM. However, the good results are achieved by jointly training with 3D object detection, which relies on a costly selective search step. Differently, our method with a single segmentation task outperforms WyPR to a large margin. Lastly, the MIL-derived Transformer~\cite{yang2022mil} modeling the similarities across scenes to improve the weak labels. However, the cross-scene similarity might be hard to model due to the large cross-scene variations. Moreover, our method can be a complementary to MIL-derived Transformer by exploiting similarities from different perspectives. 

\subsection{Ablation Study and Further Analysis}

We report the results of ablation study to show the effectiveness of each components of our method. We also conduct further analysis by qualitative results. The experiments in this section are conducted on ScanNet validation set.

\vspace{-0.6em}
\paragraph{Contributions of Components.}
We report the results of ablation study to demonstrate the contribution of different components of our method. The results are shown in Table~\ref{tab:ablation} with 8 experiments denoted by ``A.\#''. The baseline method in ``A.1'' is our MIL-based segmentation model as described in Section~\ref{sec:baseline}, which consists of a MIL-loss, a self-training loss, and a cross transformation consistency loss between the original point cloud and the geometrically augmented point cloud. The baseline method achieves 23.8\% mIoU. Then we add the multi-modality affinity to the baseline model in ``A.3'', which achieves obtains 28.4\% with 4.6\% performance gain compared to the baseline. To alleviate the imbalance issue that may affect both the affinity matrix and the segmentation results, the classifier normalization is added and the performance achieves 36.2\% as shown in ``A.4'', which significantly improve the results. ``A.8'' is our final results by using the proposed MMA-refined model, which further improve ``A.4'' by using additional multi-scale information and achieves the best results. The results also show that though simply
introducing NCW (baseline + NCW in ``A.2'') can improve
the baseline result (in ``A.1'') by 3\%, ``A.8'' (resp., ``A.4'') can further improve ``A.6'' (resp., ``A.3'') by nearly 8\%, which verifies that NCW not only balances the classification results but also enhances the point affinity. In other words, NCW only works significantly well when jointly working with the proposed MMA module. Even without NCW, we add multi-modality affinity and multi-scale affinity to baseline in ``A.6'' can achieve 30.1\% with 6.3\% performance gain compared to the baseline. Adding multi-scale affinity to the baseline in ``A.2'' can improve baseline result by 1.9\%. To validate the effectiveness of the additional light-weight Transformer block in our student segmentation head, we conduct experiments by removing the Transformer block as shown in ``A.7'', the performance drops from 37.7\% to 33.9\%. To sum up, all the components in our method contribute to the final results.

\begin{table}[!ht]
    \centering
    \caption{ Ablation study on ScanNet validation set.}
    \vspace{-1em}
    \label{tab:ablation}
    \resizebox{\linewidth}{!}{
    \begin{tabular}{c|cccc|c}
     \hline
         & Baseline & Multi-modality Aff. & Multi-scale Aff. & $\widehat{W}$ & results \\ \hline
        A.1 & \checkmark & ~ & ~ & ~ & 23.8 \\ 
        A.2 & \checkmark & ~ & ~ &  \checkmark &  26.9 \\ 
        A.3 & \checkmark & \checkmark & ~ & ~ & 28.4 \\
        A.4 & \checkmark & \checkmark & ~ & \checkmark & 36.2 \\
         A.5 &  \checkmark & ~ &  \checkmark & ~ &  25.7 \\ 
         A.6 &  \checkmark &  \checkmark &  \checkmark & ~ &  30.1 \\ 
        A.7 & w/o Trans. & \checkmark & \checkmark & \checkmark & 33.9 \\
        A.8 & \checkmark & \checkmark & \checkmark & \checkmark & \textbf{37.7} \\ \hline
    \end{tabular}
    }
    \vspace{-1em}
\end{table}

\begin{figure*}[!h]
    \centering 
\includegraphics[width=\linewidth]{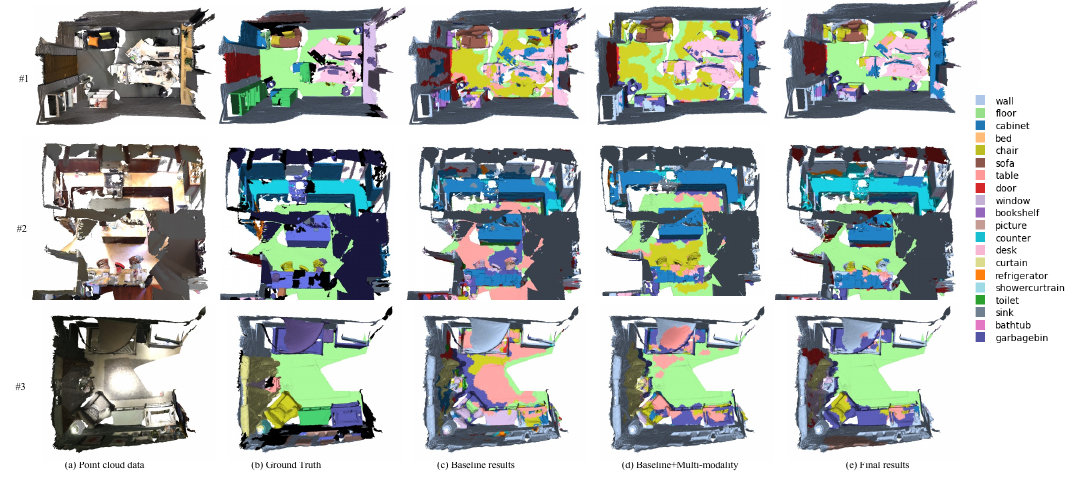}
    \vspace{-2em}
    \caption{Qualitative segmentation results: (a) Point cloud data, (b) Ground-truth segmentation results, (c) Baseline results, (d) Results of ``Baseline+Multi-modality affinity w/o $\hat{W}$'' variant of our method, (e) Results of our final method (``Baseline+Multi-modality affinity").}
    \label{fig:vis_scene_supp}
     \vspace{-1em}
\end{figure*}

\paragraph{Qualitative Results.}
We qualitatively illustrate more segmentation results in Figure \ref{fig:vis_scene_supp}, where the columns indicate:(a) Point cloud data, (b) Ground-Truth segmentation results,(c) Baseline results, (d) Results of “Baseline+Multi-modality affinity”, (e) Results of our final method. From the results,
we can find that our final method achieves the best results. 
For the baseline method (Figure \ref{fig:vis_scene_supp}(c)), different objects are not clearly distinguished with clear boundary, and the semantic categories for many objects are mis-classified.

In our final model with the additional multi-modality affinity module (Figure \ref{fig:vis_scene_supp} (e)), the discriminative information from both geometric modality and color modality can both benefit the final results. 
Although the original RGB-appended point cloud data are more informative, the RGB data and the geometric data are entangled, and thus the complementary multi-modality information is hard to be fully exploited. 
For example, in the RGB-appended point cloud data, the rich geometric data with shape and scale information might be overwhelmed by the RGB information that suffers from lighting conditions, shadows, and reflections in many cases. As shown in Figure~\ref{fig:vis_scene_supp} \#2 and \#3, ``floor'' is wrongly segmented in our single-stream baseline variant due to the serious reflections in the RGB data. By contrast, our multi-modality affinity ( Figure~\ref{fig:vis_scene_supp} (d)) successfully corrects the results thanks to the enhanced geometric information.
In addition, the ``floor'' is geometrically located with smaller ``z'' value and smaller height, which is distinguished from the objects placed on it with higher height (such as ``table'' and ``chair'') as shown in most of the scenes such as \#1, \#2, and \#3. For the ``door'' and ``wall'' classes in scene \#1, we observe that by taking advantages from the discriminative color information, our method can better distinguish ``door'' from ``wall'' when compared with the baselines.
\paragraph{Analysis of Multi-scale Affinity.}
In Table~\ref{tab:scale}, we further evaluate the effectiveness of multi-modality multi-scale affinity. Specifically, the ``$F^M_u$  only'' variant is the multi-modality affinity-only baseline. The ``$concate(F^M_u,F^M_1)$'' method concatenates the $F^M_u$ features with $F^M_1$ (i.e., features from first SA layer). The ``$concate(F^M_u,F^M_1,F^M_2)$'' method concatenates the $F^M_u$ features with both the first SA layer features $F^M_1$ and the second SA layer feature $F^M_2$. Based on multi-modality, multi-scale can further enhance the model's performance. Therefore, it is necessary for multi-modality and multi-scale to work in conjunction with each other.

\begin{table}[!ht]
    \centering
    \caption{{\bf Evaluation of the effectiveness of multi-scale affinity} on ScanNet validation set. $concate(\cdot,\cdot)$ denotes concatenation.}
    \label{tab:scale}
    \begin{tabular}{c|c}
    \hline
        Method & mIoU(\%)  \\ \hline
        $F^M_u$  & 36.2  \\ 
        $concate(F^M_u,F^M_1)$  & 36.6  \\ 
        $concate(F^M_u,F^M_1,F^M_2)$ & \textbf{37.7}  \\ \hline 
    \end{tabular}
    \vspace{-1em}
\end{table}

\paragraph{Analysis of Normalizing Classifier Weights for Affinity.}
To demonstrate the effectiveness of the Normalizing Classifier Weights (NCW) module on improving affinity, we conducted a comparative analysis between the MMA w/o $\widehat{W}$ affinity, MMA affinity. We split the ScanNet classes into three groups by the number of points in each category in training samples: head classes each contains over 8 million points, medium classes each has between 1.2 million and 8 million points, and tail classes with under 0.3 million points. 
We demonstrate in supplementary materials the categories included in the head, medium, and tail of ScanNet, respectively. 
We evaluate mAP and mIoU for each subset, the results are shown in Table~\ref{tab:ncw}, where $ \Delta$ represents the relative performance difference between MMA and MMA w/o $\widehat{W}$.
mAP reflects the performance of multi-label classification, which impact the quality of generated pseudo-labels. We observed that MMA w/o $\widehat{W}$ affinity can significantly enhance the performance of medium and tail classes while maintaining the effectiveness of the head set. Notably, there was a 7.2\% performance gain in the tail set. Due to the improved classification performance of the medium and tail categories, the points previously wrongly categorized as head class have been correctly classified to the medium and tail categories. Thus, the mIoU for all splits are improved. 



\begin{table}[!ht]
    \centering
    \renewcommand{\arraystretch}{1.3}
    \small
    \caption{mAP, mIoU of MMA and MMA w/o $\widehat{W}$ on ScanNet}
    \label{tab:ncw}
    \vspace{-1em}
    \resizebox{\linewidth}{!}{
    \begin{tabular}{c|c c c| c c c}
    \hline
        \multirow{2}{*}{method} & \multicolumn{3}{c|}{mAP} &\multicolumn{3}{c}{mIoU}\\ \cline{2-7}
            & Head & Medium & Tail & Head & Medium & Tail \\ \hline
        MMA w/o $\widehat{W}$  & 100.0 & 86.4 & 65.7 & 55.3 &18.8 & 30.6  \\
        MMA &  100.0 &  88.3& 72.9 & 68.6 &29.6 & 36.0\\ 
        $\Delta$ &  \textbf{0} & \textbf{1.9} & \textbf{7.2} &\textbf{13.3}  &\textbf{10.8} & \textbf{5.4} 
        \\\hline 
    \end{tabular}}
    \vspace{-1em}
\end{table}

\begin{figure}[ht]
    \centering    \includegraphics[width=\linewidth]{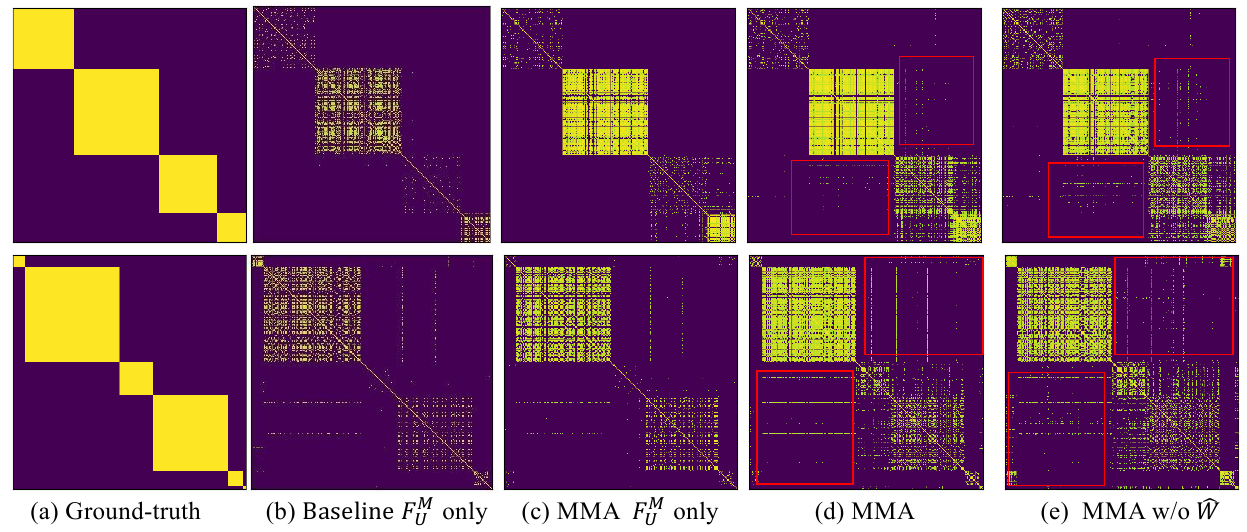}
    \vspace{-1em}
    \caption{Class relationship maps enhanced by point affinity.}
    \label{fig:multi-scale-vis}
    \vspace{-1em}
\end{figure}

\paragraph{Visualization of Point Class Relationship Maps Enhanced by Point Affinity.}
We further visualize the binary map of the class relationship enhanced by point affinity in Figure~\ref{fig:multi-scale-vis}. If two points are predicted as the same class, the value is 1, otherwise the value is 0.
The columns indicate the affinity matrices produced based on: (a) Ground-truth, (b) Baseline $F^M_u$ only,  (c) MMA $F^M_u$ only, (d) MMA (i.e., \textit{$concate(F^M_u,F^M_1,F^M_2)$}), (e) MMA w/o $\widehat{W}$. In Figure~\ref{fig:multi-scale-vis}(b), different categories get confused because the affinities between points of the same class are not higher than those between points of different classes. 
In contrast, Figure~\ref{fig:multi-scale-vis}(c) shows much better results than (b). This suggests that our MMA approach enhances the intra-class similarity by fully exploiting multiple data modalities and considering the long-tail issues. Figure~\ref{fig:multi-scale-vis}(c) and (d) demonstrate that our multi-scale features can further improve the affinity of objects from different scales. By comparing Figure~\ref{fig:multi-scale-vis}(d) and (e), the affinities of points from different classes in (d) with the proposed long-tail aware normalization is much smaller than that in (e), especially in the highlighted red boxes, which is the key to learn discriminative features.

\section{Conclusion}
This paper proposes a novel multi-modality point affinity (MMA) enhanced weakly supervised 3D semantic segmentation method. The proposed MMA considers the point feature  similarities by taking advantage of complementary information from different data modalities. The point affinity is also enhanced by normalizing the classifier weights to alleviate the detrimental effects of long-tailed data distribution to the affinity matrix. Extensive experiments demonstrate the effectiveness of the proposed method on two commonly used indoor scene understanding benchmark datasets.

\section*{Acknowledgments}
This work was supported by the National Key Research and Development Program of China (No. 2021YFB1714300), National Natural Science Foundation of China (No.62006012, No.62132001, No.62002012), in part by the Hong Kong Research Grants Council General Research Fund (17203023), in part by The Hong Kong Jockey Club Charities Trust under Grant 2022-0174,  in part by the Startup Funding and the Seed Funding for Basic Research for New Staff from The University of Hong Kong, and in part by the funding from UBTECH Robotics.

\bibliography{aaai24}

\end{document}